\documentclass{article}

\usepackage{arxiv}

\usepackage[utf8]{inputenc} 
\usepackage[T1]{fontenc}    
\usepackage{hyperref}       
\usepackage{url}            
\usepackage{booktabs}       
\usepackage{amsfonts}       
\usepackage{nicefrac}       
\usepackage{microtype}      
\usepackage{lipsum}		
\usepackage{graphicx}
\usepackage{natbib}
\usepackage{doi}
\usepackage{algorithm}
\usepackage{algpseudocode}
\newcounter{algorithmcounter}

\title{A template for the \emph{arxiv} style}

\author{ \href{https://orcid.org/0000-0000-0000-0000}{\includegraphics[scale=0.06]{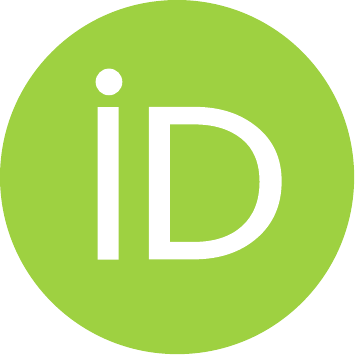}\hspace{1mm}Ali Karkehabadi} \\
	Department of Computer Electrical Engineering and Computer Science\\
	University of Davis, California\\
	California, 95615 \\
	\texttt{akarkehabadi@ucdavis.edu} \\
	\And
}

\renewcommand{\shorttitle}{\textit{arXiv} Template}

\hypersetup{
pdftitle={A template for the arxiv style},
pdfsubject={q-bio.NC, q-bio.QM},
pdfauthor={David S.~Hippocampus, Elias D.~Striatum},
pdfkeywords={First keyword, Second keyword, More},
}

\begin{document}

\title{ Does Saliency-Based Training bring Robustness for Deep Neural Networks in Image Classification?}

\maketitle
\thispagestyle{empty}
\begin{abstract}
   Deep Neural Networks are powerful tools to understand complex patterns and making decisions. However, their black-box nature impedes a complete understanding of their inner workings. While online saliency-guided training methods try to highlight the prominent features in the model's output to alleviate this problem, it is still ambiguous if the visually explainable features align with robustness of the model against adversarial examples. In this paper, we investigate the saliency trained model's vulnerability to adversarial examples methods. Models are trained using an online saliency-guided training method and evaluated against popular algorithms of adversarial examples. We quantify the robustness and conclude that despite the well-explained visualizations in the model's output, the salient models suffer from the lower performance against adversarial examples attacks. 
\end{abstract}

\section{Introduction}
\subsection{Saliency Guided Training}
Deep Neural Networks have gained immense attention from researchers in different complicated applications \cite{lecun2015deep, farabet2012learning, mikolov2010recurrent, mittal2020applications, brants2007large}. However, their black-box nature hampers their application in critical domains.  
To this end, providing explanations of the inner working process in the models has been investigated by researchers \cite{khakzar2022explanations}. Some works investigate the post-hoc explanations from the model by either evaluating backpropagated gradients \cite{selvaraju2017grad, omeiza2019smooth} or by evaluating the model's performance in the presence of a perturbation occurring in an area of the input \cite{fong2019understanding, fong2017interpretable, samek2016evaluating}. There are some works trying to shallow down the Deep Neural Network architecture to reduce the ambiguity of the model\cite{frosst2017distilling, wu2018beyond}. Others try to include the saliency in the training process.\cite{ross2017right} uses annotations on the irrelevant parts of input and penalizes the model for negative gradients to teach the model learn salient features. The authors add a regularization term that jointly optimizes the accuracy and penalization of negative gradients. \cite{ghaeini2019saliency} extends the work by involving intermediate layers, too. They teach the model to learn human-understandable features by providing ground truth masks for the positive gradient as explanations. The method trains the model with training data, labels and ground truth explanation masks.  \cite{ismail2021improving} improves the model's salient features by removing input pixels related to lower gradient values from backpropagation. Input areas with lower gradient impact are masked with random values. Authors add Kullback–Leibler (KL) divergence as a similarity metric in the loss function. The model is enforced to output similar distribution for both the original image and the masked one. The strategy helps the model to focus on the salient features which are considered to be more interpretable and be negligent to low gradient values.
\subsection{Adversarial Examples}
Although Deep Neural Networks have achieved state of the art performance in different domains, they might be vulnerable to small hardly perceptible artifact manipulation in the input. \cite{szegedy2013intriguing} Revealed that nuisance perturbations in the input images cause a misclassification by the model. Since this early study, others started developing different approaches to add a small perturbation toward the input to fool the model. \cite{goodfellow2014explaining} Introduced Fast Gradient Sign Method (FGSM). The method adds a small perturbation to all the input image pixels in the direction of the gradient. This simple but effective method built the basic block for more sophisticated attacks. \cite{kurakin2018adversarial} and \cite{madry2017towards} developed the algorithm in an iterative approach and introduced Basic Iterative Method (BIM) and Projected Gradient Descent (PGD). These approaches put a constraint on the perturbation to ensure that it is small enough to be imperceptible. \cite{dong2018boosting} considered generating strong adversarial examples using the previous momentum of gradients iteratively.  \\ Adversarial examples initially intended to attach the models for malfunction. Meanwhile, they are considered as the evaluation metrics for the model's performance against manipulated inputs. 
\\
Inspired by Saliency guided training procedures which makes the model explicitly show important features and adversarial example progresses, this paper investigates the effect of saliency-based training on the adversarial robustness of Deep Neural Networks. Although similar works like \cite{etmann2019connection} show that adversarial training for the model's robustness leads to improvement in saliency maps, we are unaware of research investigating the effect of saliency, training on adversarial robustness. This paper follows several experiments and gives insights into the connection between saliency training and robustness.  
\section{Methodology}
Our objective is to establish a fair comparison between regular and saliency-guided models against different adversarial attacks. Models are trained from scratch for classification task. Saliency-guided models are trained following \cite{ismail2021improving}. To quantify the robustness, we considered white-box attacks. White-box adversarial examples are generated with full awareness of the model's weights. We evaluated various perturbation-based algorithms following \cite{rauber2017foolbox}. We briefly review the adversarial attack algorithms in this section and discuss the results in the next sections.

\begin{algorithm}
\refstepcounter{algorithmcounter}

\caption{Saliency Training}
\begin{algorithmic}[1]
    \State \textit{Initialize $f_{\theta_0}$ }
    \For{$i = 1$ to $N_{epochs}$}
        \State \textit{Indexes = Sort($\nabla_xf_{\theta_i}$)}
        \State \textit{Masking($X(Indexes)$)}
        \State \textit{Calculating the loss: }
        
$L_i = L(f_{\theta}(X_i), y_i) + \lambda D_{KL}(f_{\theta}(X_i) | f_{\theta}(\widetilde{X}_i))$
        \State \textit{Updating the model's weights}
    \EndFor
    \State \textbf{end}
\label{alg: salient_training}
\end{algorithmic}
\end{algorithm}

\subsection{Saliency Guided Training}
Following \cite{ismail2021improving}, the saliency-guided training model is trained by adding a regularization term in the cost function. In each iteration, input images are fed to the model, gradients are calculated and sorted by their magnitude. $K$ is defined as a parameter which determines the number of the lower gradients being replaced. By attributing the gradient values to the input, related pixels are masked. Algorithm 1 shows the training procedure. Masking values are randomly selected from the uniform distribution of the original image pixel value range. To make the model learn salient features, the model's weights are updated to output same output distribution for the original and masked input. Kullback–Leibler (KL) divergence regularization term in Eq.\ref{eq: loss} guarantees this. 
\begin{equation}
\sum_{i=1}^{n} \left[L(f_{\theta}(X_i), y_i) + \lambda D_{KL}(f_{\theta}(X_i) | f_{\theta}(\widetilde{X}_i))\right] \label{eq: loss}
\end{equation}
where $X_i$ is input data, $\widetilde{X}_i$ is masked input and $\theta$ is model's parameters. Hyperparameter $\lambda$ works as a tradeoff between loss function terms.
\subsection{Fast Gradient Sign Method}
FGSM \cite{goodfellow2014explaining} is a white-box method that tries to manipulate the whole input data based on the model's weights in order to lead the model to an incorrect prediction. The perturbation occurs in the direction of the sign of the gradient. It is a  fast one-step gradient with a fixed step size. Eq.\ref{eq:einstein} shows the update algorithm where $\epsilon$ is the amount of perturbation and $y_{true}$ is the correct label for the input $x$.

\begin{equation}
    x^{\mathrm{adv}} = x + \epsilon \cdot \mathrm{sign}(\nabla_x J(x, y_{\mathrm{true}}))
    \label{eq:einstein}
\end{equation}

\subsection{Basic Iterative Method}
BIM \cite{kurakin2018adversarial} extends FGSM by applying a small amount of perturbation in each iteration. After each step, the generated input is clipped so that the final results would be in $\epsilon$-neighbourhood of the main input:

\begin{equation}
    x_0^{{adv}} = x, 
    \\
    x_{n+1}^{adv} = Clip_{x,\epsilon}(x_N^{{adv}} + \alpha{sign}(\nabla_x J(x_N^{{adv}}, y_{{true}})))
\end{equation}

\subsection{Projected Gradient Descent}
Similar to BIM, PGD \cite{madry2017towards} generates the adversarial examples in an iterative way. In each stage, the input is manipulated by solving a constrained optimization problem. The predefined amount of constraint keeps the content as the original input so that the generated example seems imperceptibly different to human eyes. The PGD is different from BIM as the initialization is set randomly.


\subsection{Momentum Iterative Method}
MIM \cite{madry2017towards} iterates over the cumulative direction of previous gradients to perturb the input.

\begin{equation}
    g_{(N+1)} = \mu g_N + \frac{\nabla_x L(f(x_N^{{adv}}, \theta^*), y)}{\|\nabla_x L(f(x_N^{{adv}}, \theta^*), y)\|_2}, \\
    x_{(N+1)}^{{adv}} = x_N^{{adv}} + \alpha \cdot {sign}(g_{(N+1)})
\end{equation}

\section{Implementation Details}
\subsection{Datasets}
\subsubsection*{MNIST} The dataset \cite{lecun2010mnist} includes 10 classes of handwritten digits containing 70000 grayscale images with the $28*28$ pixel size. We split the dataset into 60000 images for training and 10000 for test set.

\subsubsection*{CIFAR-10}Including 60000 low-resolution RGB images. The dataset contains 10 classes consisting of Airplanes, Cars, Birds, Cats, Deer, Dogs, Frogs, Horses, Ships, and trucks. Training and test set include 50000 and 10000 images respectively.
\subsubsection*{Model Architecture}
We focused on Convolution-based architectures for implementation. Our Architecutre follows \cite{ismail2021improving} for the Saliency-guided training model. For MNIST dataset, we employed a two-layer CNN network with kernel size 3 and stride of 1. Two Fully-Connected layers were followed by the network. Two drop-out layers also were used with $p=0.25$ and $p=0.5$.\\
For CIFAR-10, following \cite{ismail2021improving}, we used ResNet18 (Pretrained on ImageNet) \cite{he2016deep} and 10 neuron classifier as the output. We trained our models using single NVIDIA A100 GPU. The models were trained for 100 epochs with the batch size 256. Models used Adadelta \cite{zeiler2012adadelta} as the optimizer with learning rate $0.1$.

During the training and within each epoch, low gradient values are replaced with random values. The values are in the range of the remaining pixels for the image. Gradient values are calculated using Captum library \cite{kokhlikyan2020captum}. A fixed amount of $50\%$ of lower gradient values are removed.

\begin{table}
  \begin{center}
    {\small{
\begin{tabular}{lllllr}
\toprule
Dataset  & \# Test & \# Classes & Accuracy & SAL-Accuracy\\
\midrule
MNIST  & 10000 & 10 & 99.3\% & 98.8\%\\
CIFAR-10  & 10000 & 10& 82.8\% & 82.2\%\\
\bottomrule
\end{tabular}
}}
\end{center}
\caption{Traditional and Saliency Model Training Accuracy}
\end{table}


accuracy and for a fair comparison, we employed the same architecture for regular training. 

\subsubsection*{Robustness}
For robustness evaluation we used Foolbox \cite{rauber2017foolbox} library for adversarial attack algorithms. We evaluated the model's test accuracy for different degrees of perturbation.

\section{Results and Discussions}
\begin{figure}[t]
\begin{center}
   \includegraphics[width=0.4\linewidth]{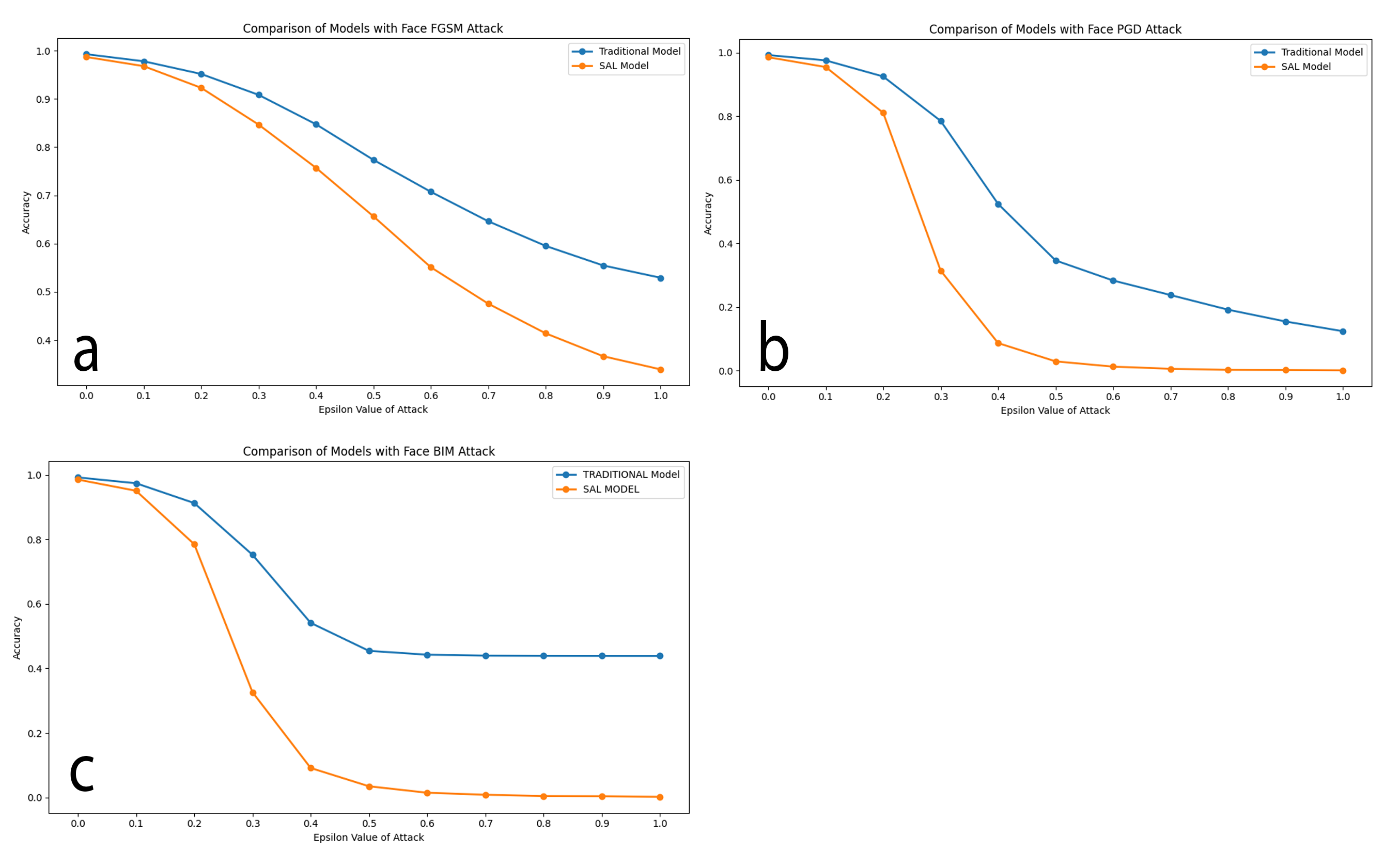}
\end{center}
   \caption{Saliency guided trained and regular model performance against attacks over MNIST dataset for different amount of perturbation. a. FGSM attack b. PGD attack c. BIM attack}

\label{fig:mnist}
\end{figure}

\begin{figure}[ht]
\begin{center}
   \includegraphics[width=0.4\linewidth]{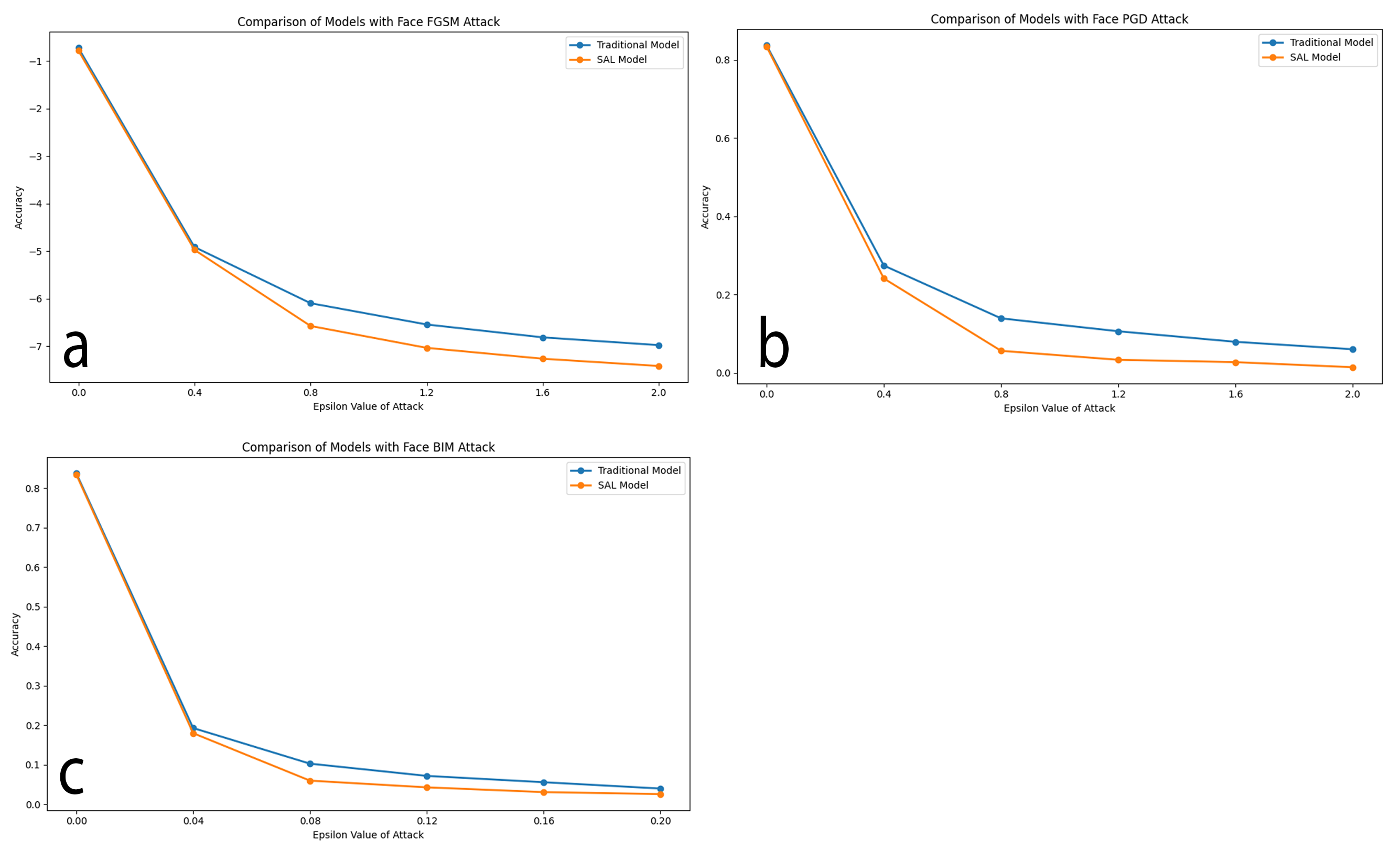}
\end{center}
   \caption{Saliency guided trained and regular model performance against attacks over CIFAR10 dataset for different amount of perturbation. a. FGSM attack b. PGD attack c. BIM attack}

\label{fig:cifar}
\end{figure}

Our purpose is to quantify the connections between robustness and saliency training approach. To this end, we trained the models following saliency-guided training \cite{ismail2021improving}. Also, we train the same architecture  without guided training as a baseline for comparison. Afterward, we evaluate the classification accuracy of the trained models against adversarial examples. Results for MNIST and CIFAR-10 are shown in \ref{fig:mnist} and \ref{fig:cifar} respectively.

\begin{figure}[ht]
\begin{center}
   \includegraphics[width=0.2\linewidth]{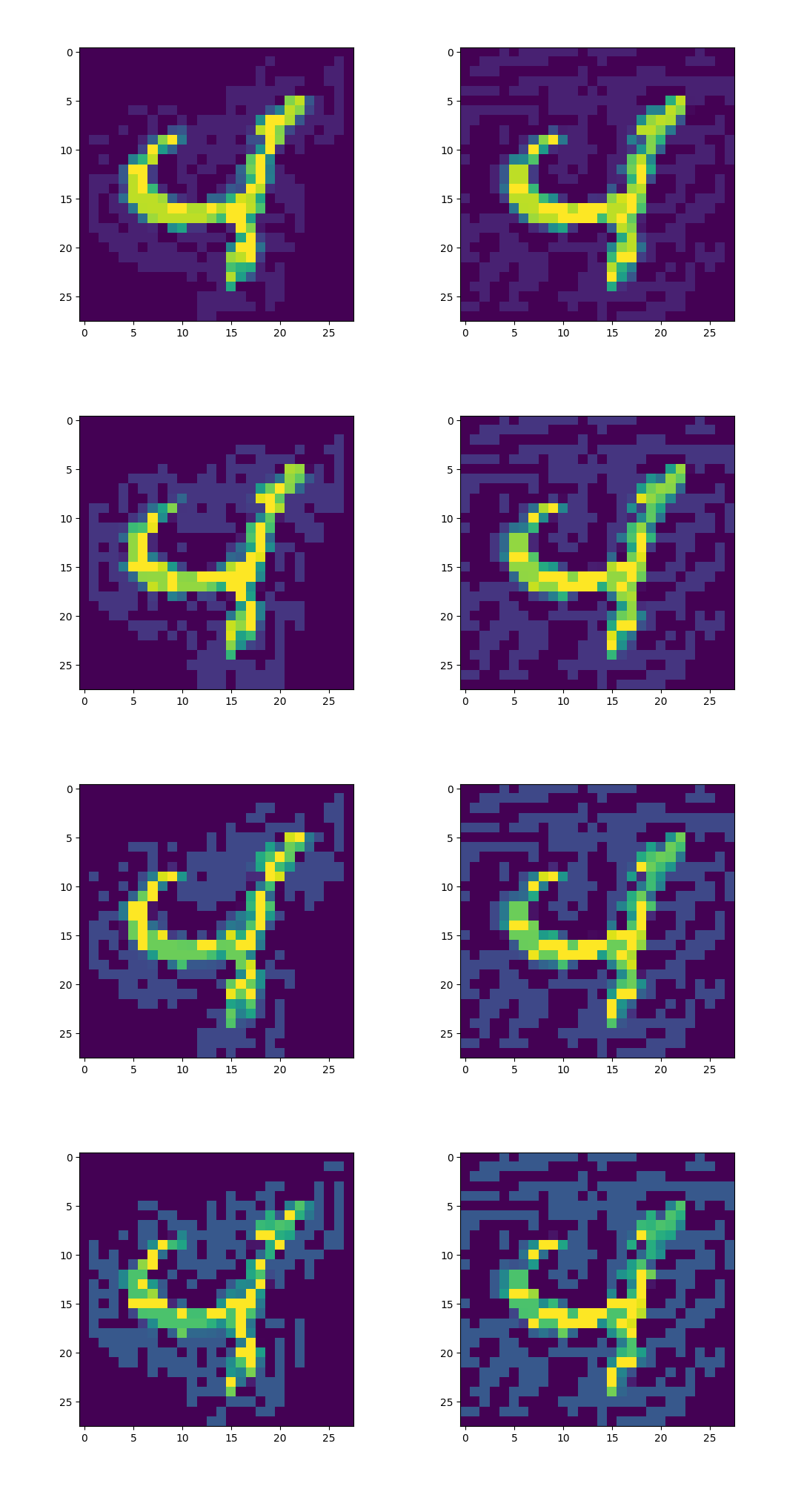}
\end{center}
   \caption{Generated FGSM Adversarial examples for Saliency guided trained (Left) and Regular model trained (Right). $\epsilon$ increases downward. } 
 \label{fig:comp2}
\end{figure}
\subsection{Conclusion}
The interpretability of Deep Learning models is of high importance based on the application. As the topic grows popular among researchers, we investigate a different aspect of salient training. Despite the fact that high values of gradient might help for better visual interpretability, our results show that it might compromise vulnerability against attacks. Through visual observations (Fig.\ref{fig:comp2}) from the models with the salient approach in their learning process, it could be figured out that by highlighting high-value gradients and their pertinent pixels that are considered as the main object, the main object is the center of attention for adversarial examples. According to the adversarial examples formula which work in the direction of gradient, results seems to be grounded. This gives way to adversarial methods to affect the model's performance.
Our work encourages and attracts the attention of the researchers to consider both aspects of saliency and robustness.  


\bibliographystyle{unsrtnat}
\bibliography{references}

\end{document}